\pdfoutput=1
\documentclass[11pt]{article}
\usepackage{latexsym}
\usepackage{EMNLP2023}

\usepackage{times}
\usepackage{latexsym}
\usepackage{booktabs} 
\usepackage{CJKutf8}
\usepackage{tabularx}
\usepackage{amsmath}
\usepackage{subcaption} 
\usepackage{multirow}
\usepackage{amsmath}
\usepackage{algorithm}
\usepackage{algpseudocode}
\usepackage{xcolor} 
\usepackage{mdframed} 
\usepackage{lipsum} 
\definecolor{lightgray}{gray}{0.9}
\usepackage{listings}
\usepackage{graphicx}
\usepackage{amssymb}
\usepackage{booktabs}
\usepackage{colortbl}
\usepackage{xcolor}
    
\usepackage{makecell}
\usepackage{amssymb} 
\usepackage[utf8]{inputenc}
\usepackage{caption}  
\usepackage{float}
\usepackage{stfloats}
\usepackage{microtype}
\usepackage{hyperref}
\usepackage{tcolorbox}

\usepackage{tikz}
\usepackage{geometry}
\newtcolorbox{mybox}{
  colback=red!10, 
  colframe=red!75!black, 
  boxrule=1pt, 
  arc=4pt, 
  boxsep=0pt, 
  left=6pt, 
  right=6pt, 
  top=6pt, 
  bottom=6pt, 
  width=\textwidth 
}

\newtcolorbox{mybox1}{
  colback=gray!10, 
  colframe=gray!75!black, 
  boxrule=1pt, 
  arc=4pt, 
  boxsep=0pt, 
  left=6pt, 
  right=6pt, 
  top=6pt, 
  bottom=6pt, 
  width=\textwidth 
}

\usepackage[T1]{fontenc}

\usepackage[utf8]{inputenc}

\usepackage{microtype}

\usepackage{inconsolata}

\title{LanguaShrink:Reducing Token Overhead with Psycholinguistics}

\makeatletter
\renewcommand{\@author}{
  \small
  Xuechen Liang \thanks{* Equal contribution} \\
  \small
  East China Jiaotong University \\
  \small
  \texttt{lxc974464857@outlook.com}
  \And
  \small
  Meiling Tao* \\
  \small
  Guangdong University of Technology \\
  \small
  \texttt{3221010067@mail2.gdut.edu.cn}
  \And
  \small
  Yinghui Xia* \\
  \small
  AutoAgents.ai \\
  \small
  \texttt{vix@autoagents.ai}
  \AND
  \small
  Tianyu Shi \\
  \small
  University of Toronto \\
  \small
  \texttt{tianyu.s@outlook.com}
  \And
  \small
  Jun Wang \\
  \small
  East China Normal University \\
  \small
  \texttt{wongjun@gmail.com}
  \And
  \small
  Jingsong Yang\thanks{ ~ Corresponding author: \texttt{edward.yang@autoagents.ai}} \\
  \small
  AutoAgents.ai \\
  \small
  \texttt{edward.yang@autoagents.ai}
}
\makeatother

\begin{document}
\maketitle
\begin{abstract}
As large language models (LLMs) improve their capabilities in handling complex tasks, the issues of computational cost and efficiency due to long prompts are becoming increasingly prominent. To accelerate model inference and reduce costs, we propose an innovative prompt compression framework called LanguaShrink. Inspired by the observation that LLM performance depends on the density and position of key information in the input prompts, LanguaShrink leverages psycholinguistic principles and the Ebbinghaus memory curve to achieve task-agnostic prompt compression. This effectively reduces prompt length while preserving essential information. We referred to the training method of OpenChat.The framework introduces part-of-speech priority compression and data distillation techniques, using smaller models to learn compression targets and employing a KL-regularized reinforcement learning strategy for training.\cite{wang2023openchat} Additionally, we adopt a chunk-based compression algorithm to achieve adjustable compression rates. We evaluate our method on multiple datasets, including LongBench, ZeroScrolls, Arxiv Articles, and a newly constructed novel test set. Experimental results show that LanguaShrink maintains semantic similarity while achieving up to 26 times compression. Compared to existing prompt compression methods, LanguaShrink improves end-to-end latency by 1.43 times.
\end{abstract}

\begin{figure}[t]
    \centering
    \includegraphics[width=\linewidth]{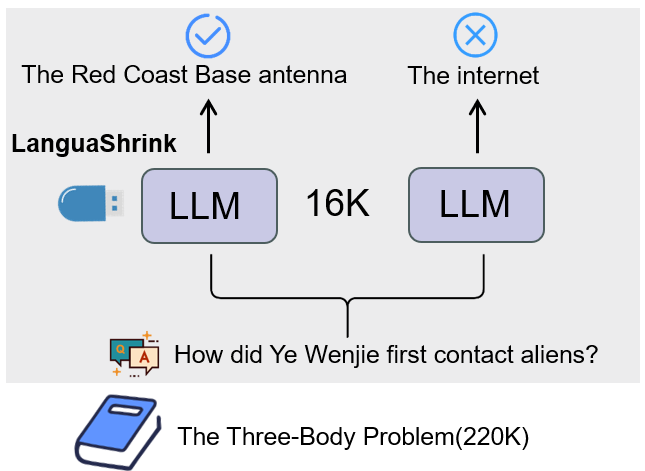}
    \caption{Illustration of the Plug-and-Play Document Module. The document encoding is decoupled from specific tasks. By inserting the document plugin into the task model, we can separate compressed text from downstream task reasoning and reduce computational costs.}
    \label{fig1}
\end{figure}

\section{Introduction}

In recent years, the field of large language models (LLM) has seen the emergence of various prompting techniques, such as Chain of Thought (CoT)\cite{Wei_Wang_Schuurmans_Bosma_Chi_Le_Zhou}, In-context Learning (ICL)\cite{dong2022survey}, and Retrieval Augmented Generation (RAG)\cite{Lewis_Perez_Piktus_Petroni_Karpukhin_Goyal_Küttler_Lewis_Yih_Rocktäschel_etal_2020}. These techniques have greatly expanded the capabilities of LLMs in handling complex and diverse tasks, by using prompts that can contain up to tens of thousands of vocabulary tokens.\cite{Manathunga2023Aligning} However, while such lengthy prompts enhance processing capabilities, they also bring higher computational costs and financial burdens, posing challenges to the information processing and comprehension abilities of LLMs.\cite{Zhou2023Adaptive-Solver}

To alleviate these issues, prompt compression techniques have emerged, aiming to reduce the length of the original prompts while preserving the core information and key instructions as much as possible, in order to optimize costs and efficiency.\cite{Mu2023Learning} Currently, many methods have been proposed for task-specific prompt compression, but these methods lack generality and portability. On the other hand, some other studies have explored task-agnostic prompt compression methods to pursue better generality and efficiency. These methods assume that natural language contains redundant information\cite{Jiang2023LongLLMLingua:}, which may be useful for human understanding but might not be necessary for LLMs.

However, current task-agnostic methods face several challenges. Existing compression techniques mainly rely on simple token classification, which may lead to the loss of important sentence structure information\cite{Kuvshinova2019Improving}. For complex long-text processing, effectively compressing without sacrificing the inherent logic and semantic structure of sentences remains an inadequately addressed issue\cite{Wang2019Improving}. Additionally, most existing models do not effectively evaluate the importance of each sentence within a paragraph, which is crucial for maintaining the coherence and completeness of information in long texts.\cite{Luo2022A}

To address these issues, we propose a new framework based on psycholinguistics, called LanguaShrink.LanguaShrink combines plug-and-play modules and psycholinguistic models to parse document information, using the Ebbinghaus memory curve to filter important information. This enables task-agnostic prompt compression and adapts to various open-source and proprietary large models. As shown in Figure \ref{fig1}, LanguaShrink can decouple compressed texts from downstream task reasoning and reduce computational costs. \cite{Hu2013Emulating,Murre2015Replication}

Specifically, we use plug-and-play modules for compression, segment the text into chunks, and evaluate the semantic and structural importance of each chunk to avoid losing critical information. By using a comprehensive weighting method, we assess the relevance and perplexity of the chunks, selecting those with high relevance and low perplexity to improve the coherence and completeness of the compressed text. Additionally, we propose a data distillation method that uses small models to learn the compression target, thereby reducing latency\cite{Ma2020Adversarial}. We incorporate a reinforcement learning framework based on KL regularization, refining the training process with different reward weights.

We validate the effectiveness of our method on three datasets from different domains, namely Longbench\cite{bai2023longbench}, ZeroScrolls\cite{shaham-etal-2023-zeroscrolls}, and Arxiv Articles\cite{clement2019arxiv}, and we also construct a new long-text novel test set. Experimental results show that our method achieves better semantic similarity compared to existing prompt compression methods at the same compression rate, 
while reducing end-to-end latency by 1.43 times and achieving a compression ratio of 2x to 8x.

The main contributions of our work are as follows:
\begin{itemize}
    \item We propose a plug-and-play prompt compression system based on psycholinguistics and the Ebbinghaus memory curve to filter important information.
    \item We propose a data distillation method that uses smaller models to learn the compression target, optimizing training through a reinforcement learning framework based on KL regularization.
    \item We conduct extensive experiments on various datasets, and the results demonstrate that our method achieves up to 26x compression without compromising performance.
\end{itemize}

\begin{figure*}[t]
    \centering
    \includegraphics[width=\linewidth]{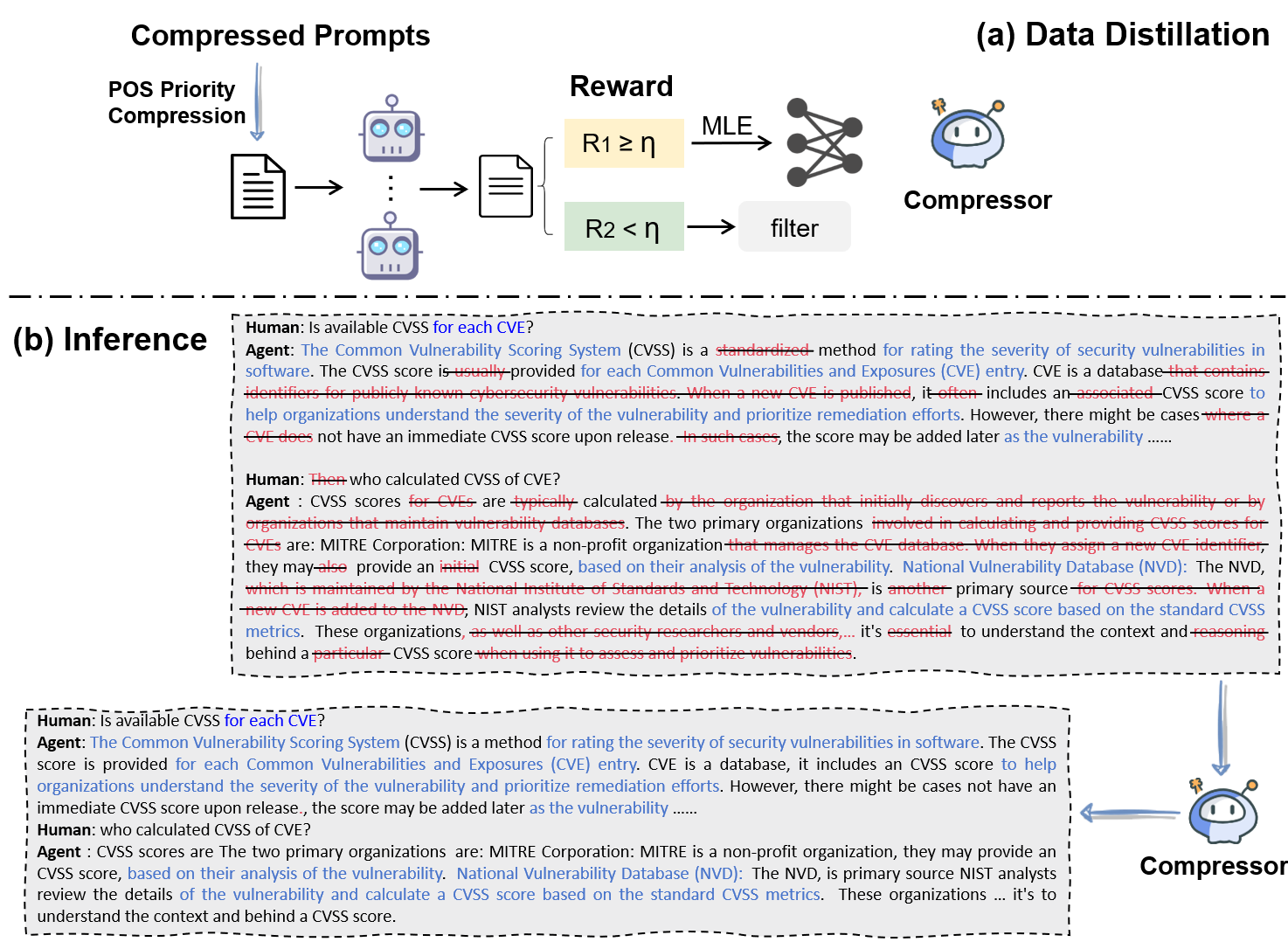}
    \caption{(a) Data distillation. Initial text compression is first performed using POS priority compression. Next, the compressed prompts are evaluated based on the similarity and compression ratio between the compressed prompt and the original prompt. If the similarity is above the threshold, the model receives a reward; otherwise, the reward is zero and it is filtered out. Then, the model is fine-tuned using Maximum Likelihood Estimation (MLE), and finally, the compressor generates the compressed prompts. (b) Inference. The application of the compressor in actual question-answering tasks is demonstrated. The effect of LinguaShrink compression processing on the original dialogue is shown. \textcolor{red}{Red} indicates the parts that are most likely to be compressed, \textcolor{blue}{blue} indicates the parts that are next most likely to be compressed.
}
    \label{fig:1}
\end{figure*}

\section{Related work}

\subsection{Psycholinguistics}

Psycholinguistic research has two main areas: sentence processing and text processing\cite{McKoon1998Memory-based}. Sentence processing focuses on how the syntactic structure of sentences is computed. \cite{Alyahya2018Noun}
Text processing involves understanding the meaning of larger units of text. Conversely, function words and key nouns play crucial roles in sentences
\cite{Kalyuga2012Cognitive}

Existing research indicates that removing redundant information can effectively improve the efficiency of foreign language vocabulary learning\cite{Ellis1993Psycholinguistic}. 
Additionally, this helps optimize storage space\cite{Schmidhuber2000Neural}. Based on these research findings, we propose a psycholinguistics-based Part-of-Speech Priority Compression (PPC) algorithm that uses lexical classification and priority assignment to more efficiently retain core information and eliminate redundant content\cite{Graça2011Controlling}.

\subsection{Prompt Compression}
LLMs face significant challenges in handling long contexts. Due to the quadratic growth in memory and computational demands of the attention mechanism, the computational cost of processing long texts is extremely high\cite{Han2023HyperAttention:,Zhuang2022Long-range,Chen2023LongLoRA:}. Existing LLMs typically use a fixed context window during pre-training, which further limits their ability to handle longer contexts. To address this issue, researchers have proposed methods such as sparse attention and local dense attention to reduce computational and memory costs. Additionally, soft prompt tuning and reinforcement learning-based compression methods have been applied to save context costs during inference.\cite{Shen2018Reinforced,Liu2023Deja}

Prompt compression is one of the direct methods to address the problem of LLMs handling long contexts, aiming to shorten prompt length while retaining important information. Typical prompt compression methods are divided into task-aware and task-agnostic categories. Task-aware compression adjusts the context based on downstream tasks or current queries, such as LongLLMLingua\cite{Jiang2023LongLLMLingua:}, which adjusts the compression ratio by estimating token information entropy. Task-agnostic compression, on the other hand, applies to a wide range of applications and typically uses information entropy measures to remove redundant information from the prompt. Although these methods have improved computational efficiency and model performance to some extent, further exploration and optimization are needed to effectively handle long texts and complex tasks in a wide range of real-world applications\cite{Hsieh2023Automatic}.

The OpenChat project proposed an effective mechanism for prompt compression fine-tuning. This method dynamically adjusts prompts through reinforcement learning strategies to retain the most important information for tasks, thereby reducing unnecessary computational overhead. Additionally, the Kullback-Leibler (KL) divergence is used to measure changes in prompt information before and after compression. By minimizing KL divergence, it ensures that critical information from the original prompt is preserved during compression.

In this paper, we propose the Prompt Compression Fine-Tuning (PC-RLFT) method, which achieves efficient prompt compression and core information retention by incorporating chunk compression techniques.

\section{Method}

\subsection{POS Priority Compression}
PPC is achieved by inputting carefully designed prompts into LLMs. To implement PPC, we need to design a series of specific prompts that achieve part-of-speech priority compression through the CoT approach. Below is the design idea for CoT:

\noindent\textbf{Relation Word Extraction }
We design prompts to guide the model in identifying relation words in the text. By using dependency syntax analysis, the model can understand the relationships between sentences. The model assigns different priorities to each relation word based on the context of the entire sentence.

\noindent\textbf{Part-of-Speech Classification }
The prompts guide the model to classify words in the text by their parts of speech, such as adjectives, adverbs, nouns, and prepositions. This is based on psycholinguistic part-of-speech analysis, assigning a priority to each part of speech. For example, the priority standard is: nouns > verbs > adjectives > adverbs.

\noindent\textbf{Priority Filtering }
After completing the extraction of relation words and part-of-speech classification, the model uses this information to filter out the words and sentences that contribute most to the core meaning of the text, while deleting lower-priority words and sentences that have a minimal impact on the overall understanding.



\begin{algorithm}
\caption{Compression Algorithm}
\begin{algorithmic}[htbp]
\Require{$T$: input text}

\State $C \gets SplitToChunks(T)$
\State $n \gets ChunkCount(C)$

\For{$i \gets 1$ to $n$}
    \State $c_i \gets C[i]$
    \State $TokenCompression(c_i)$ 
\EndFor

\State $T' \gets JoinChunks(C)$ 

\State \Return $T'$

\end{algorithmic}
\end{algorithm}

\subsection{Dataset Distillation}
 We propose a data distillation method that extracts knowledge from large language models (LLMs) to generate compressed prompts that retain key information while reducing latency by using smaller models to learn the compression targets. Additionally, we ensure the compressed prompts remain highly faithful to the original content.

\textbf{Dataset:} We source our data from reading materials in the Chinese Gaokao and the postgraduate entrance exam English sections. These reading materials provide a rich variety of texts suitable for compression training. The dataset contains 20,000 samples, each processed by segmenting the reading materials into blocks of three consecutive sentences. This structure allows the model to learn effective compression while avoiding the loss of critical contextual information.

\begin{algorithm}[H]
\caption{Chunk-Based Compression Algorithm}
\begin{algorithmic}[htbp]
\Require{$T$: input text, $Q$: query, $\alpha$: relevance weight, $\beta$: importance weight, $ R_t $: target compression rate}

\State $T' \gets Compression(T)$ 
\State $C \gets SplitToChunks(T')$ 
\State $n \gets ChunkCount(C)$
\State $R_0 \gets CalcCompRate(T, T')$ 

\For{$i \gets 1$ to $n$}
    \State $c_i \gets C[i]$
    \State $rel_i \gets CosineSim(c_i, Q)$ 
    \State $imp_i \gets CalcImportance(c_i)$ 
    \State $ppl_i \gets CalcPerplexity(c_i)$ 
    \State $w_i \gets \alpha \times rel_i + \beta \times imp_i$ 
    \State $C[i] \gets (c_i, w_i, ppl_i)$ 
\EndFor

\State $SortByWeightAndPerp(C)$ 

\State $k \gets n$
\State $R \gets R_0$

\While{$R > R_t $}
    \State $T'' \gets JoinTopKChunks(C, k)$ 
    \State $R \gets CalcCompRate(T, T'')$ 
    
    \If{$R \leq R_t $}
        \State \textbf{break}
    \EndIf
    
    \State $k \gets k - 1$ 
\EndWhile

\State \Return $T'', R$

\end{algorithmic}
\end{algorithm}

To generate compressed data, we use various LLMs, including both open-source and proprietary models such as GPT-4\footnote{https://openai.com/}, Yi\footnote{https://www.lingyiwanwu.com/}, GLM\footnote{https://chatglm.cn/}, and Qwen\footnote{https://tongyi.aliyun.com/qianwen/}. These LLMs are used to distill knowledge according to psycholinguistic principles, creating compressed prompts that retain subject-verb-object structures and maximize the preservation of key information. We also ensure the compressed content remains semantically similar to the original content. The selection of multiple LLMs is based on their complementary strengths in language understanding and generation. By integrating knowledge from multiple models, we achieve a more comprehensive compression perspective, enhancing the quality of the compressed data and the model's generalization capabilities.

\subsection{Prompt Compress-RLFT}

\subsubsection{Reward Design}
The reward consists of two components: one is based on the cosine similarity score, which measures the similarity between the output sequences generated from the original and compressed prompts; the other is the compression ratio $\tau$, reflecting the reduction in prompt length. If the cosine similarity score exceeds a certain threshold $\tau$, the model receives the compression ratio as a reward; if it does not, the reward is zero.

\subsubsection{Tuning}

We selected the pre-trained Qwen as the smaller language model (SLM). The distilled dataset is then used for Prompt Compress Reinforcement Learning Fine Tuning (PC-RLFT). During fine-tuning, we combine the PC-RLFT method based on a KL regularization reinforcement learning framework. We assign different reward weights to the data to refine the training process.

KL-regularized RL objective is defined as follow:
\begin{equation}
J_{PC-RLFT}(\theta) = \mathbb{E}_{y\sim\pi_\theta}\left[ r_c(x, y) \right] - \beta D_{KL}(\pi_\theta, \pi_c)
\label{equ:1}
\end{equation}

 where \( \pi_\theta \) is the policy parameterized by \( \theta \), \( r_c(x, y) \) is the class-conditioned reward function, \( \pi_c \) is the higher-quality class-conditioned behavior policy, \( \beta \) is a scaling factor for the KL divergence term, and \( D_{KL} \) represents the KL divergence.\cite{wang2023openchat}

Previous work has demonstrated that the optimal solution to the KL-regularized reward maximization objective is as follows:
\begin{equation}
\pi^*(y|x, c) \propto \pi_c(y|x, c) \exp\left(\frac{1}{\beta} r_c(x, y)\right)
\label{equ:2}
\end{equation}

where \( \pi^* \) signifies the optimal policy for a given class \( c \) and input \( x \).

The method to extract the optimized policy \( \pi_\theta \) by minimizing the KL divergence:
\begin{equation}
\begin{split}
\pi_\theta = \arg\min_\theta \mathbb{E}_{(x,c)\sim D_c} [D_{KL}(\pi^*(\cdot|x, c) \| \pi_\theta(\cdot|x, c))] \\
= \max_\theta \mathbb{E}_{(x,y,c)\sim D_c} \left[\exp\left(\frac{1}{\beta} r_c(x, y)\right) \log \pi_\theta(y|x, c)\right]
\end{split}
\label{equ:3}
\end{equation}

Equation~\ref{equ:3} outlines the process for minimizing the KL divergence between \( \pi^* \) and \( \pi_\theta \) over the class-conditioned dataset \( D_c \). The final expression represents the reward-weighted regression objective for the optimized policy \( \pi_\theta \).







    
    



In this study, we propose a chunk-based compression algorithm. First, the input text \( T \) is preprocessed through a standard token compression process, segmenting it into chunks \( C[i] \) consisting of three consecutive sentences. This step achieves a high initial compression rate \( R_0 \).

\subsection{Chunk-Based Compression}

For each chunk \( C[i] \), the algorithm evaluates its relevance to the query \( Q \) by calculating the cosine similarity \( \text{CS}(C[i], Q) \). The relevance is denoted as \( \text{rel}_i \). The semantic and structural importance of the chunk is calculated using the function \( \text{CI}(C[i]) \), producing an importance score \( \text{imp}_i \). Additionally, the perplexity \( \text{ppl}_i \) of the chunk, as a measure of information content, is calculated using the function \( \text{CP}(C[i]) \).

The combined weight \( w_i \) of each chunk is calculated using the following formula:
\[ w_i = \alpha \times \text{rel}_i + \beta \times \text{imp}_i \]
where \( \alpha \) and \( \beta \) are coefficients that adjust the influence of relevance and importance. This weight determines the retention priority of the chunk in the final compressed text.

The chunks are then sorted based on their weight \( w_i \) and perplexity \( \text{ppl}_i \), with higher weights and lower perplexities being prioritized for retention to optimize information preservation and compression effectiveness. The number of retained chunks \( k \) is adjusted in a decremental manner until the compression rate \( R \) reaches the target compression rate \( R_t \). Finally, the selected chunks are recombined to form the compressed text.

\begin{table*}[t]
\begin{minipage}{\textwidth}
\centering
\resizebox{\textwidth}{!}{
\begin{tabular}{lccccccccccccc}
\toprule
\multicolumn{2}{c}{} & \multicolumn{8}{c}{LongBench} & \multicolumn{4}{r}{ZeroSCROLLS} \\
\cmidrule(r){3-11} \cmidrule(r){12-14}
Methods & & SingleDoc & MultiDoc & Summ. & FewShot & Synth. & Code & AVG & Tokens & 1/$\tau$ & AVG & Tokens & 1/$\tau$ \\
\midrule
\multicolumn{14}{c}{2,000 tokens constraint} \\
\cmidrule(r){3-14}
\multicolumn{2}{l}{\textbf{Retrieval-based Methods}} \\
BM25 & & 30.1 & 29.3 & 21.3 & 12.5 & 19.5 & 29.1 & 23.63 & 1802 & 5x & 20.1 & 1,799 & 5x  \\
SBERT & & 33.8 & 36.0 & 25.8 & 23.5 & 12.5 & 29.0 & 23.6 & 1947 & 5x & 20.5 & 1,773 & 5x \\
OpenAI & & 34.3 & 36.4 & 24.6 & 26.3 & 32.4 & 24.8 & 30.47 & 1991 & 5x & 20.6 & 1,784 & 5x \\
LongLLMLingua  & & \underline{37.8} & \underline{41.7} & \textbf{26.9} & 64.3 & \textbf{53.0} & 52.4 & \underline{46.0} & 1960 & 5x & 24.9 & 1,771 & 5x \\
\midrule
\multicolumn{2}{l}{\textbf{Compression-based Methods}} \\
Selective-Context\cite{li2023unlocking} & & 16.2 & 34.8 & 24.4 & 8.4 & 15.7 & 49.2 & 24.8 & 1925 & 5x & 19.4 & 1,865 & 5x \\
LLMLingua & & 22.4 & 32.1 & 24.5 & 61.2 & 10.4  & 56.8 & 34.6 & 1,950 & 5x & 27.2 & 1,862 & 5x \\
LLMLingua2-small & & 29.5 & 32.0 & 24.5 & \underline{64.8} & 22.3 & 56.2 & 38.2 & 1,891 & 5x & \underline{33.3} & 1,862 & 5x \\
LLMLingua2 & & 29.8 & 33.1 & 25.3 & \textbf{66.4} & 21.3  & \textbf{58.9} & 39.1 & 1,954 & 5x & \underline{33.3} & 1,898 & 5x \\
\midrule
LanguaShrink & & \textbf{42.1} & \textbf{54.3} & \underline{26.3}  & 62.3 & \underline{33.0} & \underline{58.4} & \textbf{46.1} & 1,988 & 5x & \textbf{39.0} & 1,871 & 5x \\
\midrule
\multicolumn{14}{c}{3,000 tokens constraint} \\
\cmidrule(r){3-14}
\multicolumn{2}{l}{\textbf{Retrieval-based Methods}} \\
BM25 & & 32.3 & 34.3 & 25.3 & 57.9 & 45.1 & 48.9 & 40.6 & 3,417 & 3x & 19.8 & 3,379 & 3x  \\
SBERT & & 35.3 & 37.4 & 26.7 & 63.4 & \textbf{51.0} & 34.5 & 41.4 & 3,399 & 3x & 24.0 & 3,340 & 3x \\
OpenAI & & 34.5 & 38.6 & \underline{26.8} & 63.4 & 49.6 & 37.6 & 41.7 & 3,421 & 3x & 22.4 & 3,362 & 3x \\
LongLLMLingua  & & \underline{37.6} & \underline{42.9} & \textbf{26.9} & \underline{68.2} & \underline{49.9} & 53.4 & \underline{46.5} & 3,424 & 3x & 33.5 & 3,206 & 3x \\
\midrule
\multicolumn{2}{l}{\textbf{Compression-based Methods}} \\
Selective-Context & & 23.3 & 39.2 & 25.0 & 23.8 & 27.5 & 53.1 & 32.0 & 3,328 & 3x & 20.7 & 3,460 & 3x \\
LLMLingua & & 31.8 & 37.5 & 26.2 & 67.2 & 8.3 & 53.2 & 37.4 & 3,421 & 3x & 30.7 & 3,366 & 3x \\
LLMLingua2-small & & 35.5 & 38.1 & 26.2 & 67.5 & 23.9 & \underline{60.0} & 41.9 & 3,278 & 3x & 33.4 & 3,089 & 3x \\
LLMLingua2 & & 35.5 & 38.7 & 26.3 & \textbf{69.6} & 21.4 & \textbf{62.8} & 42.4 & 3,392 & 3x & \underline{35.5} & 3,206 & 3x \\
\midrule
LanguaShrink & & \textbf{42.2} & \textbf{54.5} & 26.3 & 62.6 & 34.0  & \textbf{62.8} & \textbf{47.1} & 3,488 & 3x & \textbf{39.6} & 3,197 & 3x \\
\midrule
Original Prompt & & 41.7 & 38.7 & 26.5 & 67.0 & 37.8  & 54.2 & 44.9 & 10,295 & - & 34.7 & 9,788 & - \\
\bottomrule
\end{tabular}
}
\caption{Performance of different methods under different compression ratios on LongBench and ZeroSCROLLS using GPT-3.5-Turbo.}
\label{tab:performance_comparison}
\end{minipage}
\end{table*}

\section{Experiment}

\subsection{Settings}

\textbf{Implementation Details}  We use the PPC method to analyze the text, extracting a large compressed dataset. By training the model using the PC-RLFT approach, we obtain a smaller model. All reported metrics use GPT-3.5 as the target LLM for downstream tasks. To improve the stability of the outputs generated by the LLM, we apply greedy decoding with a temperature of 0 in all experiments. In our experiments, we utilize Qwen-1.8B as the smaller pre-trained language model for compression.

\textbf{Dataset and Metrics}  To comprehensively evaluate the effectiveness of compressed prompts in retaining LLM capabilities, we assess their performance across multiple datasets. For long-context scenarios, we use LongBench and ZeroSCROLLS.

\begin{enumerate}
    \item[i.] \textbf{LongBench:\cite{bai2023longbench}} This benchmark consists of six task types: single-document QA, multi-document QA, summarization, few-shot learning, code completion, and synthetic tasks. We evaluate using the English portion, covering 16 datasets. We use the provided metrics and scripts for evaluation.

    \item[ii.]  \textbf{ZeroSCROLLS:\cite{shaham-etal-2023-zeroscrolls}} This benchmark comprises four task types: summarization, QA, sentiment classification, and reordering, covering 10 datasets. We evaluate using the validation set and employ the provided metrics and scripts for evaluation.

    \item[iii.]  \textbf{Arxiv-March:\cite{clement2019arxiv}} A dataset composed of the latest academic papers covering various scientific disciplines. Due to the potential length of arXiv articles, in our experiments, we only process the first two sections of each paper (usually the introduction and background). We compress the content of all the papers and compare the effects before and after compression.

    \item[iv.]  \textbf{Novel Test:} We select a novel with nearly 250K context. We test the novel on Summarisation and Question Answering (QA). The Summarisation task aims to evaluate whether selective context affects the model's overall understanding of the input context. The Question Answering task aims to assess the model's understanding of specific queries. We compare compression time, compression quality, similarity to the original text, and end-to-end time on these tasks. Additionally, we propose the Compression Semantic Efficiency (CSE) metric, calculated through the compression ratio and similarity.
\end{enumerate}

\textbf{Baselines}  
We adopt two state-of-the-art prompt compression methods as the primary baselines for comparison: Selective-Context and the llmlingua\cite{Jiang_Wu_Lin_Yang_Qiu,Jiang2023LongLLMLingua:,Pan_Wu_Jiang_Xia_Luo_Zhang_Lin_Rühle_Yang_Lin_etal_2020} series. Additionally, we compare our method with several task-aware prompt compression methods, such as retrieval-based methods and longllmlingua.

\begin{table}
\centering
\begin{tabular}{lcccc}
\toprule
 Method &  CSE &  BLEU &  ROUGE \\
\midrule
Select context & 3.1080 & 0.0010 & 0.2063 \\
llmlingua2 & 1.4845 & 0.0008 & 0.2015  \\
LanguaShrink & 3.6555 & 0.0235 & 0.2015  \\
\bottomrule
\end{tabular}
\caption{Statistics of Arxiv Articles Cosine, BLEU, ROUGE, Tokens are averaged per document.}
\label{tab:Arxiv}
\end{table}

\begin{table}
\centering
\begin{tabular}{lcc}
\toprule
{Method} & {Tokens (avg)} & {Time (avg)} \\
\midrule
{\small \textit{LanguaShrink} } & 3502.75 & 24.29 \\
{\small \textit{LanguaShrink} (w/o psy.)} & 3811.3 & 33.99 \\
{\small \textit{LanguaShrink} (w/o SA)} & 3770.5 & 35.74 \\
\bottomrule
\end{tabular}
\caption{Method performance statistics. Tokens and time are averaged values.}
\label{tab:model-performance}
\end{table}

\subsection{Main Results}

Table \ref{tab:performance_comparison} presents the performance of various methods under different compression constraints. Despite our compression model being much smaller than LLAMa-2-7B\footnote{https://github.com/Meta-Llama/llama}  or other models used as baselines, our approach achieves better performance in both QA and synthesis tasks. Compared to the original prompts, our compressed prompts achieve comparable performance at a lower cost. Our model exhibits superior performance compared to other task-agnostic baselines, demonstrating the effectiveness of our constructed dataset and highlighting the importance and benefits of optimizing compression models using prompt compression knowledge.

Compression-based methods, such as selective context and LLMLingua, perform poorly on most tasks. This is due to their purely information entropy-based compression mechanism, which includes too much noise in the compressed results. Retrieval-based methods rely on finding the most relevant fragments to the query from a large number of documents. However, in practical applications, these fragments may contain a lot of redundant information, leading to lower overall information density.

We find that compressing tokens leads to a decline in mathematical capabilities, possibly because psycholinguistics is less sensitive to mathematical content. In the Longbench test, LLMLingua2 has a slight advantage in few-shot and code tasks, but LanguaShrink performs better in text or Q\&A compression tasks. This difference indicates that while psycholinguistic techniques have significant advantages in text compression, they are still inadequate in handling mathematical content.

In the ArXiv tests, as shown in Table \ref{tab:Arxiv}, our method also performs outstandingly in Compression Semantic Efficiency (CSE), a new metric that combines compression ratio and semantic similarity. Our results are 2.46 times better than llmlingua2. In terms of BLEU scores, our method shows significant improvement compared to Select context and llmlingua2. Regarding ROUGE scores, since our method alters sentence structures to maximize semantic retention, our performance in this metric is comparable to other methods, without significant improvement.

\begin{table}[ht]
\centering
\begin{tabular}{l c}
\toprule
 Method &  F1 \\
\midrule
llmlingua2 & 21.7 \\
Select context & 18.3 \\
LanguaShrink & \textbf{26.0} \\
LanguaShrink(w/o psy.) & 17.3 \\
LanguaShrink(w/o SA.) & 22.1 \\
original & 27.6 \\

\bottomrule
\end{tabular}
\caption{F1 Scores of Different Methods}
\label{tab:f1}
\end{table}

\begin{table}
\centering
\begin{tabular}{lccc}
\toprule
{Method} & {Tokens (avg)} & {CSE} & {BLEU} \\
\midrule
llmlingua2 & 280.10 & 0.9419 & 0.0288 \\
our & 253.15 & 1.0462 & 0.0304 \\
Select context & 270.15 & 0.9413 & 0.0273 \\
original & 343.65 & - & - \\
\bottomrule
\end{tabular}
\caption{Method performance statistics. Tokens, CSE, and BLEU are averaged values.}
\label{tab5}
\end{table}

\subsection{Ablation Study}

Our method consists of two core components: a psycholinguistic analysis module and a sentence analysis module. As shown in Table \ref{tab:model-performance} , when we remove the psycholinguistic core compression component, we find a nearly 10\% decrease in compression capability. This is mainly due to the lack of the linguistic analysis part, which leads to an inability to quickly and accurately locate tokens. When we remove the sentence analysis component, although the ability to quickly and accurately locate tokens is regained, the compression performance and efficiency decrease due to the inability to identify the key parts of sentences to compress.

In more practical scenario tests, due to the strategy of prioritizing semantic retention, LanguaShrink can only achieve a 90\% compression rate in standard mode. To achieve a higher compression rate, we propose a performance mode, which retains nearly 20\% of semantic information even at a 96\% compression rate.

As shown in Table \ref{tab:f1}, we first use the constructed novel dataset as the original context to generate questions and answers, where these answers are considered reference answers, and then require the LLM to answer these questions. We find that with the psycholinguistic core compression component present, even without the sentence analysis module, LanguaShrink can reach the level of llmlingua2.

The CSE metric we propose measures the true effectiveness of large model compression methods. When the value is below 1, compressing the large model does not improve token performance; instead, it significantly reduces token performance. When the value exceeds 1, the compression method can increase the context length. In tests with text that is already highly refined, compression may lead to a decrease in CSE performance. As shown in Table \ref{tab5}, in our tests of three models, none achieve the specified compression ratio, with the compression rate being around 30\%.  However, our method still exceeds 1 when other methods are below 1, proving the effectiveness of our method even in extreme conditions.

For more detailed cases, please go to Appendix


\begin{table}[ht]
\centering
\begin{tabular}{l c c}
\toprule
 Method & Latency(s) & Speedup Factor \\
\midrule
llmlingua & 7.48 & 1.6x \\
Select context & 7.56 & 1.6x \\
LanguaShrink &\textbf{6.64} & \textbf{1.8x} \\
original & 11.84 & - \\
\bottomrule
\end{tabular}
\caption{Latency and Speedup Factor of Different Methods}
\label{Latency}
\end{table}

\subsection{Latency Evaluation}

We conducted tests on the A800-80GB GPU, using the same prompt as indicated in the appendix, which on average contained 10K tokens, and set the response length to 200 tokens in the API calls. In Table \ref{Latency}, "E2E" represents the latency of each prompt compression system and the black-box API. The results show that our prompt compression system indeed accelerates the overall inference. This acceleration effect becomes more pronounced with the increase in compression rates. It is worth noting that in scenarios where the API's cost time is longer, the actual absolute time saved by LanguaShrink may be more significant.
\cite{Cao2023GPU-HADVPPM,Stone2008Accelerating,Yazdanbakhsh2015Neural}

\section{Conclusion}

In this paper, we propose LanguaShrink, an innovative prompt compression framework aimed at improving the efficiency and performance of LLMs by reducing the length of prompts while preserving core information. LanguaShrink leverages psycholinguistic models and the Ebbinghaus memory curve to achieve task-agnostic compression compatible with various LLMs. We introduce a method based on part-of-speech priority compression and data distillation techniques, using smaller models to learn compression targets and employing a KL-regularized reinforcement learning strategy for training. Additionally, we adopt a chunk-based compression algorithm, evaluating each chunk's relevance, importance, and perplexity to adjust the retention priority and achieve adjustable compression rates. Extensive experimental results show that LanguaShrink significantly outperforms existing techniques in semantic similarity and compression efficiency across multiple datasets, achieving up to 26 times compression while maintaining performance comparable to the original prompts.

\section*{Limitations}
Currently, our token compression technology mainly incorporates psycholinguistic techniques and has not yet integrated RAG (Retrieval-Augmented Generation) technology. In early experiments, we tried various psycholinguistic knowledge and ultimately selected the two most effective methods for experimentation, but we did not fully apply all the psycholinguistic knowledge. In these early experiments, we partially used the Oxford Dictionary for training. Although this yielded good results, we were unable to conduct comprehensive testing due to not having collected the complete content of the Oxford Dictionary.

LanguaShrink becomes unstable when the compression rate exceeds 90\%. Although we introduce an extreme mode to address this issue, it is not an ideal long-term solution. While the extreme mode can temporarily mitigate the performance degradation caused by excessive compression, it may introduce other complexities and resource consumption in practical applications.


In the future, our research focuses on further optimizing the application of psycholinguistic techniques, exploring more diverse integration methods, and addressing the decline in mathematical capabilities to achieve breakthroughs in a broader range of application scenarios.

 \section*{Ethics Statement}

The development and application of LanguaShrink also raise several ethical considerations:
Bias and Fairness: The datasets used for training and evaluating LanguaShrink must be carefully curated to ensure they are representative and do not perpetuate biases. Any inherent biases in the data could be amplified through the compression process, leading to unfair or biased outputs from the LLMs.
Privacy and Confidentiality: When applying LanguaShrink to sensitive or confidential information, it is crucial to ensure that the compression process does not inadvertently expose or compromise any personal or sensitive data. Robust data handling and privacy-preserving techniques must be implemented.
Transparency and Accountability: The use of LanguaShrink should be transparent, with clear documentation on how the compression is performed and its potential impacts on the data. Users should be informed about the limitations and potential risks associated with the compressed prompts to make informed decisions about their use.
Impact on Employment: The efficiency gains from using LanguaShrink could lead to reduced demand for certain roles involved in manual data processing and prompt generation. It is essential to consider the socio-economic impacts and provide support for individuals who might be affected by such technological advancements.

\bibliography{anthology,custom}
\bibliographystyle{acl_natbib}

\appendix

\section{Dataset Details}
\label{app:a}
Arxiv-March23 A dataset consisting of latest academic papers created in March 2023 from the arXiv preprint repository. We use 500 data items collected by Li as the test set. Due to the excessive length of some articles, we take the first five sections of each article and truncate each section to 10,000 characters. Then, we concatenate these sections to form the original prompt and use GPT-3.5-Turbo to generate the summary as the reference.

LongBench A multi-task long context benchmark consists of 3,750 problems in English and includes six categories with a total of 16 tasks. These tasks encompass key long-text application scenarios, such as single-document QA, multi-document QA, summarization, few-shot learning, synthetic tasks, and code completion. The average prompt token length in this benchmark is 10,289.

ZeroSCROLLS The multi-task long context benchmark consists of 4,378 problems, including four categories with a total of 10 tasks. These tasks cover summarization, question answering, aggregated sentiment classification, and information reordering. The average prompt token length in this benchmark is 9,788.

\section{Other implementation details}\label{app:b}

The experiment used two types of machines, but each experiment was tested on the same machine, using the A800-80GB and 3090ti. We used tiktoken11 and the GPT-3.5-Turbo model to calculate all tokens. We open-sourced an early version of the system preset instructions, which are part of the core intermediate process: Here is a set of rules for simplifying text, aimed at helping users choose the appropriate level of simplification according to different reading and comprehension needs.

\section{Rules}\label{app:c}

\subsection{Basic Rules}\label{app:c1}

\begin{enumerate}
    \item \textbf{Remove Non-Essential Information:}
    \begin{itemize}
        \item If a sentence contains two commas or dashes, consider removing the part between them unless it contains essential information.
        \item Remove all non-essential adjectives and adverbs.
    \end{itemize}
    
    \item \textbf{Simplify Clauses and Modifiers:}
    \begin{itemize}
        \item If there is a restrictive clause following a single comma, consider removing that clause.
        \item Remove all non-essential attributive, adverbial, and appositive clauses.
    \end{itemize}
\end{enumerate}

\subsection{Advanced Rules}\label{app:c2}

\begin{enumerate}
    \item \textbf{Handle Complex Relationship Sentences:}
    \begin{itemize}
        \item \textbf{Contrasting Relationships:} Retain the main information after the contrast.
        \item \textbf{Concessive Relationships:} Retain the crucial part according to contextual importance.
        \item \textbf{Causal Relationships:} Retain the reason explanation.
        \item \textbf{Result Relationships:} Highlight the factors leading to the result.
        \item \textbf{Conditional Relationships:} Retain the condition explanation.
        \item \textbf{Progressive Relationships:} Emphasize the information in the progressive part.
        \item \textbf{Comparative Relationships:} Highlight the main content of the comparison.
        \item \textbf{Coordinate Relationships:} Maintain equal treatment of content.
    \end{itemize}
    
    \item \textbf{Optional Retention:}
    \begin{itemize}
        \item Pay special attention to retaining important information such as names, places, and proper nouns.
    \end{itemize}
\end{enumerate}

\subsection{Simplification Levels}\label{app:c3}

\begin{enumerate}
    \item \textbf{Very Light Simplification:} Only remove redundant modifiers.
    \item \textbf{Light Simplification:} Apply basic comma and clause removal rules.
    \item \textbf{Moderate Simplification:} Apply all basic rules.
    \item \textbf{Deep Simplification:} Apply both basic and advanced rules, retaining key sentence meaning.
    \item \textbf{Very Deep Simplification:} Extremely reduce details, retaining only the main parts of the sentence (subject, verb, object).
\end{enumerate}

\clearpage
\section{Different compression modes}\label{app:d}

\begin{mybox}
\par
\medskip
 \textbf{Original Sentence:} "The economy, despite facing numerous challenges from external factors such as global market fluctuations and geopolitical tensions, continues to grow."
 
\textbf{Very Deep Simplification:} "The economy grow."

 \textbf{Deep Simplification:} "The economy grows despite challenges."
 
\textbf{Moderate Simplification:} "The economy grows despite external challenges."

\textbf{Light Simplification:} "The economy, despite challenges, continues to grow."

\textbf{Very Light Simplification:} "The economy, despite facing numerous challenges, continues to grow."

\medskip
\end{mybox}

\section{Cases Study}\label{app:e}
\subsection{The compression ratio of 10X.}\label{app:e1}


\begin{mybox1}
\par
\medskip
    \textbf{Original Prompt :} \\
    The author is a Reuters Breakingviews columnist. The opinions expressed are his own. NEWLINE CHAR NEWLINE CHAR BP faces opposition from some shareholders for handing Chief Executive Bob Dudley a 20 percent increase in his total remuneration package for 2015 to 19.6 million. It may seem hard to square that amount with BP’s 5.2 billion loss last year, and the fact that it is slashing thousands of jobs in response to falling oil prices. But that’s actually the point. Managing an oil company when crude is trading at 100 per barrel is easy compared to the current environment. Instead, Dudley has to work harder than his predecessors. NEWLINE CHAR NEWLINE CHAR Dudley, whose pay was going to a non-binding shareholder vote on April 14, has done what was needed of him. His two big challenges were to clean up the financial spill from the 2010 Gulf of Mexico disaster and change the culture at BP, which was tainted by safety concerns and excessive risk taking. Last year the company saw the number of recorded oil spills and employee injuries both at five-year lows. NEWLINE CHAR NEWLINE CHAR He has also delivered decent returns when compared to peers. BP ranks third among the big six oil majors, which include Exxon Mobil and Royal Dutch Shell, in total shareholder returns over the last three years, according to Eikon data – even despite 2010’s rig blowout. Drawing a line under the environmental catastrophe last year by agreeing to pay up to 18.7 billion in penalties cleared the decks for the company to start rebuilding its balance sheet. NEWLINE CHAR NEWLINE CHAR Compared to counterparts, Dudley’s remuneration appears generous. Although Shell Chief Executive Ben van Beurden pocketed 24.2 million euros (27.2 million) in 2014, this figure fell to 5.6 million euros last year, according to the company. Over the same period Dudley’s base salary has remained flat, with the biggest boost to his overall financial reward coming through his pension and deferred bonus shares. NEWLINE CHAR NEWLINE CHAR The mild-mannered American has had possibly the toughest job in the oil industry. His rewards look in line with that task.
    
    \vspace{10pt}
    \textbf{Compressed Prompt :} \\
    BP faces opposition from some shareholders for handing Chief Executive Bob Dudley a 20 percent increase in his total remuneration package for 2015 to  19.6 million.
Simplified: BP faces opposition from some shareholders for handing Chief Executive Bob Dudley a 20 percent increase in his total remuneration package for 2015.

\medskip

\end{mybox1}

\clearpage
\begin{mybox1}
\par
\medskip
    \textbf{Original Prompt :} \\
   LONDON -- A leading shareholder advisory group has criticized BP PLC's decision to award its top directors their maximum bonuses for 2015, despite the company's lackluster performance, and recommended shareholders vote against the payment plans. NEWLINE CHAR NEWLINE CHAR Last month, BP announced that Chief Executive Bob Dudley would receive a 2\% bump in his total compensation package in 2015. Though much of this increase related to U.K. reporting requirements that inflated the rise in Mr. Dudley's pension, the oil executive's cash bonus increased to  1.4 million from  1 million in 2014. His total bonus for the year, including a portion paid in deferred BP shares, amounted to  4.2 million. That was the maximum amount he was eligible to receive for the year and was up from  3 million in 2014. Chief Financial Officer Brian Gilvary also received 100\% of his possible bonus. NEWLINECHAR NEWLINE CHAR The awards follow a year in which the company lost 5.2 billion as oil prices plummeted. Since the start of 2016 it has announced plans to cut 7,000 jobs and has slashed spending to help manage the slump. NEWLINE CHAR NEWLINE CHAR ""We believe shareholders should question whether payouts were fully earned in respect of the past fiscal year relative to the company's performance,"" proxy advisory firm Glass Lewis said in a March report seen by The Wall Street Journal. NEWLINE CHAR NEWLINE CHAR BP's compensation committee awards executive bonuses based on the company's performance in a number of strategic areas, including its safety record and internal targets for operational cash flow and underlying profits. NEWLINE CHAR NEWLINE CHAR ""BP executives performed strongly in a difficult environment in 2015, managing the things they could control and for which they were accountable,"" a BP spokesman said, adding that ""safety and operational risk performance was excellent and BP responded quickly and decisively to the drop in oil price."" NEWLINE CHAR NEWLINE CHAR This isn't the first time Glass Lewis has raised objections to BP's executive pay. Last year, it also recommended that shareholders reject Mr. Dudley's pay package, noting that his compensation outpaced that received by chief executives at similar-sized firms ""despite the company's relative underperformance."" The executive's compensation was ultimately approved by around 86\% of investors. NEWLINE CHAR NEWLINE CHAR BP's shareholders will vote on the matter this year at the company's annual general meeting in London on April 16, along with a host of other issues. Glass Lewis has also raised concerns about the company's proposal to reduce its notice period for calling a general meeting, but supports most of the proposals, including the re-election of Mr. Dudley and his board. NEWLINE CHAR NEWLINE CHAR Write to Sarah Kent at sarah.kent@wsj.com NEWLINE CHAR NEWLINE CHAR More from MarketWatch

    \vspace{10pt}
    \textbf{Compressed Prompt :} \\
    A leading shareholder advisory group has criticized BP PLC's decision to award its top directors their maximum bonuses for 2015, and recommended shareholders vote against the payment plans.
Simplified: A leading shareholder advisory group has criticized BP PLC's decision to award its top directors their maximum bonuses for 2015.

\medskip

\end{mybox1}

\clearpage
\begin{mybox1}
\par
\medskip
   \textbf{Original Prompt :} \\
   Angry shareholders mounted an unprecedented protest against BP on Thursday, rebelling against a 20 per cent pay rise for chief executive Bob Dudley despite the oil group making its worst ever loss. NEWLINE CHAR NEWLINE CHAR Investors voted against the company’s pay decisions for the first time in living memory, with 59 per cent of proxy votes cast going against BP’s decision to pay Mr Dudley nearly  20m for 2015, a year in which the company ran up a  5.2bn loss. NEWLINE CHAR NEWLINE CHAR It was the first time that a top British company was defeated over executive pay since shareholders at advertising group WPP and Xstrata, the mining company, rebelled four years ago during what was dubbed the “shareholder spring”. It left BP scrambling to win back support of some of the City’s biggest institutions. NEWLINE CHAR NEWLINE CHAR The rebellion highlighted a growing trend of institutional investors and advisers around the world taking a more aggressive stance over pay. NEWLINE CHAR NEWLINE CHAR Smith  Nephew, the FTSE 100 medical devices group, also suffered a defeat on their remuneration report on Thursday as 53 per cent of shareholders voted against the pay package of chief executive Olivier Bohuon. Although Mr Bohuon’s overall pay fell to  5.5m in 2015 compared with  6.8m in 2014, shareholders protested because the company allowed long-term incentives to vest despite falling below initial targets. NEWLINE CHAR NEWLINE CHAR US banks from Citigroup to Bank of America have faced pressure to toughen bonus “clawback” regimes, which put executives on the hook for future losses. A resolution demanding more details of JPMorgan’s clawback plans attracted 44 per cent support last year. NEWLINE CHAR NEWLINE CHAR Mr Dudley’s pay looked particularly out of line to shareholders because other major energy company bosses took pay cuts in 2015, a year when energy companies were hit hard by the oil price crash. NEWLINE CHAR NEWLINE CHAR According to ISS Corporate Solutions in the US, the median pay of an S P 500 energy company chief executive, excluding their pension, fell by 1.8 per cent last year after four years of increases that ranged from 4.8 to 8.2 per cent.……(Omit here)
    
    \vspace{10pt}
    \textbf{Compressed Prompt :} \\
    Angry shareholders mounted an unprecedented protest against BP on Thursday, rebelling against a 20 per cent pay rise for chief executive Bob Dudley despite the oil group making its worst ever loss.
Simplified: Angry shareholders mounted an unprecedented protest against BP on Thursday, rebelling against a 20 per cent pay rise for chief executive Bob Dudley.

\medskip

\end{mybox1}

\clearpage
\begin{mybox1}
\par
\medskip
    \textbf{Original Prompt :} \\
   Image copyright PA Image caption Bob Dudley took over as BP chief executive in the aftermath of the fatal Gulf of Mexico oil rig explosion NEWLINE CHAR NEWLINE CHAR BP shareholders have rejected a pay package of almost £14m for chief executive Bob Dudley at the oil company's annual general meeting. NEWLINE CHAR NEWLINE CHAR Just over 59\% of investors rejected Mr Dudley's 20\% increase, one of the largest rejections to date of a corporate pay deal in the UK. NEWLINE CHAR NEWLINE CHAR The vote is non-binding on BP, but earlier, chairman Carl-Henric Svanberg promised to review future pay terms. NEWLINE CHAR NEWLINE CHAR Mr Dudley received the rise despite BP's falling profits and job cuts. NEWLINE CHAR NEWLINE CHAR Corporate governance adviser Manifest says the vote is at or above the fifth-largest in the UK against a boardroom remuneration deal. NEWLINE CHAR NEWLINE CHAR 'Last chance saloon' NEWLINE CHAR NEWLINE CHAR In his opening address to the shareholders' meeting, before the vote had been formally announced, Mr Svanberg acknowledged the strength of feeling, saying: ""Let me be clear. We hear you."" NEWLINE CHAR NEWLINE CHAR He continued: ""We will sit down with our largest shareholders to make sure we understand their concerns and return to seek your support for a renewed policy."" NEWLINE CHAR NEWLINE CHAR ""We know already from the proxies received and conversations with our institutional investors that there is real concern over the directors' pay in this challenging year for our shareholders. NEWLINE CHAR NEWLINE CHAR ""On remuneration, the shareholders' reactions are very strong. They are seeking change in the way we should approach this in the future,"" he said. NEWLINE CHAR NEWLINE CHAR The Institute of Directors said the shareholder rebellion would ""determine the future of corporate governance in the UK"". NEWLINE CHAR NEWLINE CHAR ""British boards are now in the last chance saloon, if the will of shareholders in cases like this is ignored, it will only be a matter of time before the government introduces tougher regulations on executive pay,"" said director general Simon Walker. NEWLINE CHAR NEWLINE CHAR Media playback is unsupported on your device Media caption Dudley's pay sends 'wrong message' investor says NEWLINE CHAR NEWLINE CHAR 'Out of touch' NEWLINE CHAR NEWLINE CHAR Shareholders that criticised the pay deals included Aberdeen Asset Management and Royal London Asset Management. NEWLINE CHAR NEWLINE CHAR Investor group Sharesoc branded the pay deal ""simply too high"", while Glass Lewis, ShareSoc, Pirc and Institutional Shareholder Services have also expressed their opposition. NEWLINE CHAR NEWLINE CHAR Earlier on Thursday, Ashley Hamilton Claxton, corporate governance manager at Royal London, told the BBC: ""The executives received the maximum bonuses possible in a year when [BP] made a record loss, and to us that just does not translate into very good decision-making by the board. NEWLINE CHAR NEWLINE CHAR ""We think it sends the wrong message. It shows that the board is out of touch."" NEWLINE CHAR NEWLINE CHAR She told the BBC's Today programme that if 20\%-25\% of shareholders vote down the pay deal, it would force BP to ""think long and hard about their decision"". NEWLINE CHAR NEWLINE CHAR The early voting figures suggest that the opposition is even bigger that she expected……(Omit here)
    
    \vspace{10pt}
    \textbf{Compressed Prompt :} \\
    BP shareholders have rejected a pay package of almost £14m for chief executive Bob Dudley at the oil company's annual general meeting.
Simplified: BP shareholders have rejected a pay package of almost £14m for chief executive Bob Dudley.

\medskip

\end{mybox1}

\clearpage
\begin{mybox1}
\par
\medskip
    \textbf{Original Prompt :} \\
   A majority of BP PLC’s shareholders voted against the company’s executive pay policy, a stinging — though nonbinding — rebuke to Chief Executive Bob Dudley and his board. NEWLINE CHAR NEWLINE CHAR At the company’s annual meeting Thursday, the oil giant said preliminary results showed 59 \% of investors voting by proxy rejected the company’s executive compensation decisions for 2015. That included a controversial 20 \% increase in Dudley’s total pay for the year, at a time when the company lost  5.2 billion. NEWLINE CHAR NEWLINE CHAR Earlier in the day, the company also signaled in its clearest terms yet that the oil giant may have to reduce its dividend, as low oil prices continue to threaten the once-sacrosanct investor payouts across the industry. NEWLINE CHAR NEWLINE CHAR Both moves heap pressure on Dudley and his board, as they try to navigate low oil prices like the rest of the industry but also contend with increasing shareholder unease. NEWLINE CHAR NEWLINE CHAR BP BP., +0.88 \% BP, +0.60 \% Chairman Carl-Henric Svanberg, speaking to investors before the vote, defended the pay package, which he said was based on “exceptional” company performance during a difficult year. He said, before the vote, that the board would discuss possible changes to its compensation plan for next year. NEWLINE CHAR NEWLINE CHAR After the vote, Svanberg said that despite the nonbinding vote, the company wouldn’t adjust Dudley’s pay. NEWLINE CHAR NEWLINE CHAR An expanded version of this report appears on WSJ.com NEWLINE CHAR NEWLINE CHAR More from MarketWatch
"

    \vspace{10pt}
    \textbf{Compressed Prompt :} \\
    A majority of BP PLC’s shareholders voted against the company’s executive pay policy, a stinging — though nonbinding — rebuke to Chief Executive Bob Dudley and his board.
Simplified: A majority of BP PLC’s shareholders voted against the company’s executive pay policy.

\medskip

\end{mybox1}

\begin{mybox1}
\par
\medskip
    \textbf{Original Prompt :} \\
   Item 15, report from City Manager Recommendation to adopt three resolutions. First, to join the Victory Pace program. Second, to join the California first program. And number three, consenting to to inclusion of certain properties within the jurisdiction in the California Hero program. It was emotion, motion, a second and public comment. CNN. Please cast your vote. Oh. Was your public comment? Yeah. Please come forward. I thank you, Mr. Mayor. Thank you. Members of the council. My name is Alex Mitchell. I represent the hero program. Just wanted to let you know that the hero program. Has been in California for the last three and a half years. We're in. Over 20. We're in 28 counties, and we've completed over 29,000 energy efficient projects to make homes. Greener and more energy efficient. And this includes anything. From solar to water. Efficiency. We've done. Almost.\$ 550 million in home improvements.

    \vspace{10pt}
    \textbf{Compressed Prompt :} \\
    Item 15 report City Manager Recommendation adopt three resolutions. join Victory Pace program. Second join California first program. three inclusion properties jurisdiction California Hero program. emotion motion second public comment. CNN. cast vote. public comment?. come forwar. Members council. Alex Mitchell. represent hero program. California three and a half years. 20. 28 counties completed 29,000 energy efficient projects. includes. solar water. Efficiency Almost \$550 million home improvements.

\medskip

\end{mybox1}

\clearpage
\subsection{Single-document QA in 5x constraint}\label{e2}

\begin{mybox1}
\par
\medskip
\textbf{Compressed Prompt:} \\
banned. In Roth, the Court ruled that obscenity was not protected by the First Amendment, but did not provide a clear definition of what constituted obscenity. This led to the Memoirs test, which required that material be ""utterly without redeeming social value"" to be considered obscene. However, this test was difficult to apply and led to inconsistent rulings across different jurisdictions.

The Miller decision
In Miller v. California, the Supreme Court modified its definition of obscenity, creating the three-prong Miller test. Under this test, material is considered obscene if:

The average person, applying contemporary community standards, would find that the work, taken as a whole, appeals to the prurient interest.
The work depicts or describes, in a patently offensive way, sexual conduct specifically defined by the applicable state law.
The work, taken as a whole, lacks serious literary, artistic, political, or scientific value.
This new test allowed for more local control over obscenity laws, as it required that the material be judged based on contemporary community standards. It also provided a clearer definition of what constituted obscenity, allowing for more consistent rulings across different jurisdictions.

Impact
The Miller decision had a significant impact on obscenity laws in the United States. It allowed for more local control over what was considered obscene, as it required that material be judged based on contemporary community standards. It also provided a clearer definition of obscenity, allowing for more consistent rulings across different jurisdictions. The Miller test is still used today to determine what constitutes obscenity under the First Amendment."

Question:Which case was brought to court first Miller v. California or Gates v. Collier ?

    \vspace{10pt}
    \textbf{LLMs' Response:} \\
    Miller v. California
    
    \vspace{10pt}
    \textbf{Ground Truth:} \\
    Miller v. California

\medskip

\end{mybox1}

\clearpage
\subsection{QA on LongBench with 2X constraints}\label{e3}
\begin{mybox1}
\par
\medskip
    \textbf{Question Prompt :} \\
   The Little Drummer Boy was followed by another special in 1969, Frosty the Snowman, based on the song of the same name, with narration by the American comedian Jimmy Durante and animation by Mushi Production. The success of the special led to a sequel, Frosty's Winter Wonderland, in 1976, as well as a 1979 animated adaptation of the 1957 song ""Rudolph the Red-Nosed Reindeer"" by Johnny Marks, ""Rudolph and Frosty's Christmas in July"".
In 1970, Rankin/Bass produced Santa Claus Is Comin' to Town, based on the song of the same name and starring the American actor Fred Astaire as the narrator and the voice of the mailman, Mickey Rooney as the voice of Kris Kringle/Santa Claus, and Keenan Wynn as the voice of the Winter Warlock. This was followed by The Year Without a Santa Claus in 1974, based on the 1956 book of the same name by Phyllis McGinley, with narration by the American actor Shirley Booth and the voices of Mickey Rooney as Kris Kringle/Santa Claus and Dick Shawn as the voice of Snow Miser. The success of the special led to a live-action remake in 2006, written by Larry Wilson, and a sequel, A Miser Brothers' Christmas, in 2008.
In 1977, Rankin/Bass produced The Easter Bunny is Comin' to Town, narrated by the American actor Fred Astaire and starring the voices of Skip Hinnant, Vincent Price, and Robert Morse. The studio's last major holiday special was Jack Frost in 1979, narrated by the American actor Buddy Hackett and starring the voices of Robert Morse, Debra Clinger, and Paul Frees. After Jack Frost, Rankin/Bass continued to produce animated television specials and series until 1987.
In addition to their holiday specials, Rankin/Bass also produced several non-holiday themed animated series, including The Jackson 5ive and The Osmonds, as well as the animated versions of The King Kong Show and The Smokey Bear Show. They also produced several animated feature films, including The Last Dinosaur and The Bermuda Depths.
In 1987, Rankin/Bass was acquired by Lorimar-Telepictures, which was later acquired by Warner Bros. in 1989. The studio continued to produce animated specials and series until 1994, when it was shut down. In 2001, Warner Bros. released a DVD box set of the studio's holiday specials, and in 2018, Warner Bros. announced plans to produce new films and television specials based on the studio's classic characters and properties."
Question:For what type of work is the production company for The Year Without a Santa Claus best known?

    \vspace{10pt}
    \textbf{LLMs' Response:} \\
    seasonal television specials, usually done in stop motion animation

    \vspace{10pt}
    \textbf{Ground Truth:} \\
    seasonal television specials, particularly its work in stop motion animation

\medskip

\end{mybox1}

\clearpage
\subsection{Sum.  on LongBench with 2X constraints}\label{e4}
\begin{mybox1}
\par
\medskip
    \textbf{Question: Prompt :} \\
    Project Manager: I hope you're all prepared for this feature design meeting.

Marketing Department: Of course.

Project Manager: I'll be taking notes, and the three of you will be presenting. We need to understand the requirements for the new project, including user needs, technical implementation, and design approach.

Marketing Department: I'm the marketing expert, Sammy Benjo, and I'd like to share insights on user needs and preferences for the remote control. We conducted a survey and found that users dislike the current remote's appearance and complexity of operation. They want a more aesthetically pleasing and user-friendly remote.

Project Manager: Next up.

Industrial Designer: Please press F5 for fullscreen display.

Marketing Department: I suggest we ask users about their feelings and requirements for the current remote. The survey shows users dislike the appearance and complexity of operation of the remote and hope for improvements in these aspects.

Project Manager: Alright, what's next?

Industrial Designer: The survey indicates that users' main complaints about the remote include difficulty in finding it, too many and unused buttons, among others.

Project Manager: I agree, these are all issues that need addressing.

Marketing Department: Users want the remote to have voice recognition capabilities, but there are significant differences in demand across different age groups……

   Summarize the discussion about the working design of the remote control.

    \vspace{10pt}
    \textbf{LLMs' Response:} \\
    The discussion about the working design of the remote control focused on several key aspects. Firstly, the team talked about the importance of understanding user needs, with Marketing highlighting users' preferences for simplicity, aesthetics, and functionality, such as easy zapping and a desire for fewer, more relevant buttons. The Industrial Designer discussed the technical components necessary for the remote, including a power source (battery), user interface (keypad/buttons), a digital signal processing chip for handling functions, and an infrared LED source for communication with devices. User Interface emphasized the need for a simple, intuitive design that could potentially include features like speech recognition for ease of use and automatic volume adjustment based on the user's distance from the TV. The team also considered making the remote universal for various devices, not just TVs, and discussed incorporating feedback mechanisms and possibly a screen for better interaction. The idea of speech recognition was debated, with concerns about complexity versus the potential appeal to younger users who might appreciate personalized, voice-activated controls.

    \vspace{10pt}
    \textbf{Ground Truth:} \\
    The industrial designer described the product as a portable device with some energy source, an interface which included some pressing buttons and keys, a processor, some infrared source which could send the information to the main device, and different encryption codes for different TVs. Since the Marketing and the user interface designer mentioned speech recognition, the industrial designer also took that into account. All these components would cost 5 Euros or even less than that. As for gesture recognition and automatic power control, the group gave them up due to price issues. At the end of the meeting, the group also talked about how to enable the users to find the controller.

\medskip

\end{mybox1}

\clearpage
\subsection{Sum.   on LongBench with 2X constraints}\label{e5}
\begin{mybox1}
\par
\medskip
    \textbf{Compressed Prompt :} \\
    "MOTHER, have you heard about our summer holidays yet?" said Julian, at the breakfast-table.
"Can we go to Polseath as usual?"
"I'm afraid not," said his mother. "They are quite full up this year."The three children at the breakfast-table looked at one another in great disappointment. They did solove the house at Polseath. The beach was so lovely there, too, and the bathing was fine.
"Cheer up," said Daddy. "I dare say we'll find somewhere else just as good for you. And anyway,Mother and I won't be able to go with you this year. Has Mother told you?""No!" said Anne. "Oh, Motheris it true? Can't you really come with us on our holidays? Youalways do."
"Well, this time Daddy wants me to go to Scotland with him," said Mother. "All by ourselves!
And as you are really getting big enough to look after yourselves now, we thought it would be ratherfun for you to have a holiday on your own too. But now that you can't go to Polseath, I don't reallyquite know where to send you."
"What about Quentin's?" suddenly said Daddy. Quentin was his brother, the children's uncle.
They had only seen him once, and had been rather frightened of him. He was a very tall, frowningman, a clever scientist who spent all his time studying. He lived by the sea but that was about allthat the children knew of him!
"Quentin?" said Mother, pursing up her lips. "Whatever made you think of him? I shouldn't think he'dwant the children messing about in his little house.""Well," said Daddy, "I had to see Quentin's wife in town the other day, about a business matterand I don't think things are going too well for them. Fanny said that she would be quite glad if shecould hear of one or two people to live with her for a while, to bring a little money in. Their house isby the sea, you know. It might be just the thing for the children. Fanny is very niceshe would look after them well."
"Yes and she has a child of her own too, hasn't she?" said the children's mother. "Let me seewhat's her name something funny yes, Georgina! How old would she be? About eleven, I shouldthink."
2
"Same age as me," said Dick. "Fancy having a cousin we've never seen! She must be jolly lonely allby herself. I've got Julian and Anne to play with but Georgina is just one on her own. I shouldthink she'd be glad to see us."
"Well, your Aunt Fanny said that her Georgina would love a bit of company," said Daddy. "Youknow, I really think that would solve our difficulty, if we telephone to Fanny and arrange for thechildren to go there. It would help Fanny, ……

    \vspace{10pt}
    \textbf{Compressed Prompt:} \\
   "MOTHER have you heard about our summer holidays yet?" said Julian at the breakfast-table.
"I'm afraid not," said his mother.
"Cheer up," said Daddy.
"No!" said Anne.
"Well, this time Daddy wants me to go to Scotland with him," said Mother.
"What about Quentin's?" suddenly said Daddy.
"Quentin?" said Mother.
"Well," said Daddy, "I had to see Quentin's wife in town the other day."
"Yes and she has a child of her own too, hasn't she?" said the children's mother.
"Same age as me," said Dick.
"Quentin?" said Mother, pursing up her lips.
"Well, your Aunt Fanny said that her Georgina would love a bit of company," said Daddy.
"Yes and she will love looking after you all," said Daddy.
"Well, that's settled," he said.
"Next week, if Mother can manage it," said Daddy.
"Yes," she said.
"How lovely it will be to wear shorts again," said Anne.
"Well, you'll soon be doing it," said Mother.
"Anne wanted to take all her fifteen dolls with her last year," said Dick.
"No, I wasn't," said Anne.
"Daddy, are we going by train or by car?" he asked.
"By car," said Daddy.
"That would suit me well," said Mother.
"So Tuesday it was," said Mother.
"It's a lovely day, hurrah!" cried Julian.
"It's come at last!" she said.
"Are we picnicking soon?" asked Anne.
"Yes," said Mother.
"Oh, gracious!" said Anne.
"What time shall we be at Aunt Fanny's?" asked Julian.
"About six o'clock with luck," said Daddy.
"We must watch out for the sea," said Dick.

\medskip

\end{mybox1}

\end{document}